\documentclass{article}

\usepackage[numbers]{natbib}

\usepackage[preprint]{neurips_2021}

\usepackage[utf8]{inputenc} 
\usepackage[T1]{fontenc}    
\usepackage{hyperref}       
\usepackage{subfig}
\usepackage{microtype}
\usepackage{graphicx}
\usepackage{url}            
\usepackage{booktabs}       
\usepackage{amsfonts}       
\usepackage{nicefrac}       
\usepackage{microtype}      
\usepackage{xcolor}         

\usepackage{amsthm}
\usepackage{amsmath,bm}
\usepackage{algorithm}
\usepackage{algorithmic}
\usepackage{multirow}

\def \A {\mathcal{A}}

\def \R {\mathbb{R}}

\def \R {\mathbb{R}}

\def \A {\mathcal{A}}

\DeclareMathOperator*{\argmax}{arg\,max}
\DeclareMathOperator*{\argmin}{arg\,min}
\newcommand*{\Z}{\mathbb{Z}}
\newtheorem{theorem}{Theorem}[section]
\newtheorem{definition}{Definition}

\newtheorem{lemma}[theorem]{Lemma}

\title{Deep Fair Discriminative Clustering}

\author{%
  Hongjing Zhang \\
  Department of Computer Science\\
  University of California - Davis\\
  \texttt{hjzzhang@ucdavis.edu} \\
  \And
  Ian Davidson \\
  Department of Computer Science \\
  University of California - Davis \\
  \texttt{davidson@cs.ucdavis.edu} \\
}

\begin{document}
\setlength{\abovedisplayskip}{2pt}
\setlength{\belowdisplayskip}{2pt}
\maketitle

\begin{abstract}
Deep clustering has the potential to learn a strong representation and hence better clustering performance compared to traditional clustering methods such as $k$-means and spectral clustering. However, this strong representation learning ability may make the clustering unfair by discovering surrogates for protected information which we empirically show in our experiments. In this work, we study a general notion of group-level fairness for both binary and multi-state protected status variables (PSVs). We begin by formulating the group-level fairness problem as an integer linear programming formulation whose totally unimodular constraint matrix means it can be efficiently solved via linear programming. We then show how to inject this solver into a discriminative deep clustering backbone and hence propose a refinement learning algorithm to combine the clustering goal with the fairness objective to learn fair clusters adaptively. Experimental results on real-world datasets demonstrate that our model consistently outperforms state-of-the-art fair clustering algorithms. Our framework shows promising results for novel clustering tasks including flexible fairness constraints, multi-state PSVs and predictive clustering.
\end{abstract}

\section{Introduction}
Clustering is essential as it is the basis of many AI tools and has been widely used in real-world applications \cite{jain1999data} such as market research, social network analysis, and crime analysis. However, as AI tools augment and even replace humans in decision making, particularly on other humans, the need to ensure clustering is fair becomes paramount. Here fairness is measured using protected status variables (PSVs) such as gender, race, or education level. Fairness takes two primary forms: i) group-level fairness and ii) individual-level fairness. In this paper, we study the former which ensures that no one cluster contains a disproportionately small number of individuals with protected status. Motivated by this goal, our work aims to add fairness rules to deep clustering to generate fair clusters.

Recent work \cite{chierichetti2017fair,rosner2018privacy,schmidt2019fair,kleindessner2019guarantees,backurs2019scalable,bera2019fair} has been proposed on fair clustering algorithms. To ensure group-level fairness, many of these works use the notion of the \emph{disparate impact doctrine} \cite{chierichetti2017fair}, that instances from different protected groups must have approximately (within a tolerance) equal representation in a cluster compared to the population. Different geographic regions place this tolerance at different levels. These existing fair clustering algorithms optimize the clustering quality by minimizing some well-known clustering objectives while satisfying the group-level fairness constraints. Previous examples of adding fairness to clustering algorithms include k-median based approaches \cite{chierichetti2017fair,backurs2019scalable,bera2019fair} and spectral clustering based algorithm \cite{kleindessner2019guarantees}. However, most of these algorithms evaluate their performance on low-dimensional tabular data and mainly study the problems with binary PSV. 

Deep clustering  \cite{xie2016unsupervised,hu2017learning,guo2017improved,ijcai2019-509} has the ability to simultaneously cluster and learn a representation for problems with large amounts of complex data (i.e., images, texts, graphs). However, the representation learning ability sometimes makes the learner suffer from bias hidden in the data which can lead to unfair clustering results. For example, clustering of portraits may create clusters based on features which are surrogates for racial and other protected status information. One way to overcome this is by adding group-level fairness to deep clustering which is a challenging and understudied problem. A significant challenge is it is hard to translate the current fair clustering algorithms into an end-to-end deep clustering setting.
For example, geometric pre-processing steps such as computing fairlets \cite{chierichetti2017fair} to ensure fairness will not work as the end-to-end learning of deep learners means the underlying features that clustering is performed on are unknown apriori. Similarly, another line of work that adds constraints into deep learning models such as \cite{xu2017semantic} and \cite{zhang2019deep} are not appropriate either as these constraints are at the \emph{instance} level, whereas we require to apply fairness rules at a cluster level. 

The work on fair deep clustering is relatively new. The first work on fair deep clustering \cite{wang2019towards} studies deep fair clustering problem from a geometric perspective which aims to learn a fair representation with multi-state PSV. The most recent work (and only other work) \cite{li2020deep} proposes a deep fair visual clustering model with adversarial learning to encourage the clustering partition to be statistically independent of each sensitive attribute. Although these deep clustering approaches demonstrate better clustering performance compared to the traditional fair clustering algorithms (Table \ref{tab:performance}), their fairness results are relatively poor compared to those fair clusterings with fairness guarantees \cite{chierichetti2017fair,backurs2019scalable}. Our work can be seen to combine the benefits of deep representation learning and discrete optimization by producing an intermediate clustering assignments that are guaranteed to be fair.

In this paper, we propose a novel deep fair clustering framework to address the above issues. We adopt a probabilistic discriminative clustering network and learn a representation that naturally yields compact clusters. To incorporate the group-level fairness rules in the deep learner, we first formulate our fairness objective as an integer linear programming (ILP) problem that guarantees group-level fairness. 
This ILP is efficient to solve as its constraint matrix is totally unimodular. Further, we propose a refinement learning algorithm to combine the solved fair assignments, clustering objective and self-supervised contrastive learning. Finally, we connect all these components and train the network in an end-to-end fashion. Experimental results on real-world datasets demonstrate that our model achieves near-optimal fairness with competitive clustering results. We also examine the novel uses of our framework in predictive clustering, flexible fair clustering, and challenging tasks with multi-state PSVs. 
The major contributions of this paper are summarized as follows: 
\begin{itemize}
\item Our work optimizes a general notion of fairness for multi-state PSVs which we prove (Theorem \ref{th:fair}) is equivalent to optimizing the balance measure for disparate impact \cite{chierichetti2017fair}. 
\item We formulate our fairness objective as an ILP and show the constraint matrix is totally unimodular, (Theorem \ref{thm:eff}) so it can be efficiently solved by an LP solver  (but still generate integer solutions). This allows scaling of our work.
\item We propose an end-to-end refinement learning algorithm that combines deep learning and discrete optimization that can generate meaningful and fair clusters. 
\item Extensive experimental results show that our work can achieve near-optimal fairness with competitive clustering performance. We demonstrate our novel extensions for fair clustering tasks in predictive clustering, multi-state PSVs and flexible fairness rules. 
\end{itemize}

In the next section we discuss the related work. Then we outline our measure of fairness and how it relates to classic measures of disparate impact. Next we introduce our clustering framework and encode our fairness objective as an ILP which can be efficiently solved by LP solver. A refinement learning algorithm and contrastive learning are adopted for end-to-end fair clustering. Finally we validate the effectiveness of our approach on image and tabular data.

\section{Related Work}
Fair clustering has received much attention recently \cite{schmidt2019fair,kleindessner2019fair,ahmadian2019clustering,chen2019proportionally,davidson2019making,mahabadi2020individual,brubach2020pairwise}. Chierichetti et al \cite{chierichetti2017fair} first addressed the disparate impact for clustering problems in the presence of binary PSVs. Their work apriori groups instances into many fairlets which are used as input into standard k-medians style algorithms. Their work is guaranteed to produce a specified level of fairness and achieve a constant factor approximation with respect to cluster quality. Other work \cite{backurs2019scalable} improves the fair decomposition algorithm to linear run-time. Rösner and Schmidt \cite{rosner2018privacy} generalizes the fairness notion to allow for more than two protected groups. Later on, Bera et al. \cite{bera2019fair} propose a general fair clustering algorithm that allows human-specified upper and lower bounds on any protected group in any cluster. Their work can be applied to any clustering problems under $\ell_{p}$ norms such as k-median, k-means, and k-center. Kleindessner et al. \cite{kleindessner2019guarantees} extends the fairness notion to graph spectral clustering problems.  

Previous fair clustering approaches mainly focus on adding fairness constraints into traditional clustering algorithms. In our work, we aim to study the fairness problem for recently proposed deep clustering algorithms \cite{xie2016unsupervised,yang2017towards,hu2017learning,caron2018deep,shaham2018spectralnet,tzoreff2018deep,shah2018deep}. Deep clustering algorithms connect representation learning and clustering together and have demonstrated their advantages over the two-phase clustering algorithms which use feature transformation first and then clustering. However, limited work has been done to study the fairness of those algorithms. One of the earliest works \cite{wang2019towards} to address the deep fair clustering problem learns a latent representation such that the cluster centroids are equidistant from every ``fairoid'' (the centroid of all the data belonging to the same protected group). Recently, \cite{li2020deep} encodes the fairness constraints as an adversarial loss and concatenates the fairness loss to a centroid-based deep clustering objective as a unified model. Their goal of adding fairness via adversarial loss is to make the clustering assignments to be statistically independent of the PSVs. 

Different from previous deep fair clustering work, we translate the fairness requirements into an ILP problem and our formulation allows for a general notion of fairness that supports flexible fairness constraints and multi-state PSVs. Moreover, we propose a novel learning framework to train fair clustering models via simultaneous clustering and fitting the self-generated fairness signals. Empirical studies show that our model can achieve near-optimal fairness and competitive clustering results.

\section{Definitions of Group-level Fairness}
We begin this section by overviewing the seminal definition of group-level fairness in clustering (see equation \ref{old_fair}) and then its extension to multi-state PSVs (see equation \ref{eq:old_fair_multi}). We then go onto show a new measure that our deep clustering framework will optimize (see equation \ref{new_fair}) and equation \ref{eq:old_fair_multi} have the same optimal condition as shown in Theorem \ref{th:fair}.
\subsection{Notion of Fairness}
Let $X \in {\R}^{N \times D}$ denote $N$ data points with $D$ dimension features. The prediction function $\phi$ assigns each instance to one unique cluster, $\phi : x \rightarrow \{1, ... K\}$, which forms $K$ disjoint clusters $\{C_1, ... C_K\}$. Given the protected status variable (denoted as PSV) $\A$ with $T$ states, $X$ can be partitioned into $T$ demographic groups as $\{ G_1, G_2, ... G_T\}$. 

\begin{definition}
The seminal proposed measure of fairness for clustering with binary PSV \cite{chierichetti2017fair}  encoded disparate impact as follows:
\begin{equation}
\label{old_fair}
 balance(C_k) = \min (\frac{N_{k}^1}{N_{k}^2}, \frac{N_{k}^2}{N_{k}^1}) \in [0, 1]
\end{equation}
\end{definition}
\vskip -0.1in
Here $N_{k}^1$ and $N_{k}^2$ represent the populations of the first and second demographic groups in cluster $C_k$. Such a measure of fairness only works for binary PSV. To allow for multi-state PSVs, let ${N_{k}^{min}} = \min (N_{k}^1 \ldots N_{k}^T)$ denotes the smallest (in size) protected group in cluster $k$ and ${N_{k}^{max} = \max(N_{k}^1 \ldots N_{k}^T)}$ denotes the largest group. We extend the balance measure for multi-state PSV as:
\begin{equation}
\label{eq:old_fair_multi}
 balance(C_k) = \frac{N_{k}^{min}}{N_{k}^{max}} \in [0, 1]
\end{equation}

Recent works \cite{rosner2018privacy,bera2019fair} also propose a new fairness measure to allow for fair clustering problems with multi-state PSVs. 
\begin{definition}
Let $\rho_i$ be the representation of group $G_i$ in the dataset as $\rho_i = {|G_i|} / {N} $, and $\rho_i(k)$ be the representation of group $G_i$ in the cluster $C_k$: $\rho_i(k) = {|C_k \cap G_i |} / {|C_k|}$. Using these two values, the fairness value for cluster $C_k$ is:
\begin{equation}
\small
\label{new_fair}
 fairness(C_k) = \min (\frac{\rho_i}{\rho_i(k)}, \frac{\rho_i(k)}{\rho_i}) \in [0,1] \quad\forall i \in \{1, \ldots T\}
\end{equation}
\end{definition}
The overall fairness of a clustering is defined as the \emph{minimum} fairness value over all the clusters. Similarly the overall balance is the \emph{minimum} balance value of all the clusters. 


\subsection{Equivalence of Optimizing Fairness and Balance Measures}
Here we show that optimizing equation \ref{new_fair} is equivalent to optimizing our extended definition of balance in equation \ref{eq:old_fair_multi}. 
We see that equation \ref{new_fair} achieves maximal fairness when $P(x \in G_t | x \in C_k) = \rho_t$. 
Our balance measure in equation \ref{eq:old_fair_multi} achieves optimal balance when $P(x \in G_t | x \in C_k) = \frac{1}{T}$ for any protected group $G_t$ in cluster $C_k$. 
However, this is an ideal case as protected groups may be imbalanced. Denote the size of each protected group as $|G_i|$ and the size of the data set as $N$, we now show that the optimal balance is achieved if and only if $P(x \in G_t | x \in C_k) = \rho_t$. This result indicates the equivalence of optimizing fairness (equation \ref{new_fair}) and generalized balance (equation \ref{eq:old_fair_multi}).  
\begin{lemma}
\label{lemma_balance}
The optimal balance can be achieved only when all the clusters have the same balance. Formally, $\forall i, j \in \{1, 2, ..., K\}$: $balance(C_i) = balance(C_j)$.
\end{lemma}

\begin{theorem}
To achieve optimal balance value for multi-state protected variables, we must satisfy the condition: $P(x \in G_t | x \in C_k) = \rho_t$ which is precisely the optimal fairness value for equation \ref{eq:old_fair_multi}.
\label{th:fair}
\end{theorem}

\section{Deep Fair Clustering Algorithm}

We introduce our deep fair clustering framework in this section. We can view our  algorithm as  learning a good fair clustering under three objectives: a clustering loss function, self-generated fairness signals and a contrastive learning objective. 

\subsection{Deep Clustering Model}
We learn a neural network $f_{\theta}$ as a discriminative function to predict the clustering assignments $Y = \sigma(f_{\theta}{(X)}) \in {\R}^{N \times K}$ based on input $X \in {\R}^{N \times D}$ and softmax function $\sigma$. Based on previous work \cite{krause2010discriminative,hu2017learning} we represent the mutual information $I(X;Y)$ between $X$ and clustering labels $Y$ as the difference between marginal entropy $H(Y)$ and conditional entropy $H(Y{\mid}X)$:
\begin{equation}
\begin{split}
    I(X;Y) = H(Y) - H(Y{\mid}X) 
    = h(\frac{1}{N} \sum_{i=1}^{N} \sigma(f_{\theta}(x_i))) - \frac{1}{N}\sum_{i=1}^{N} h(\sigma(f_{\theta}(x_i)))
\end{split}
\end{equation}
where $h$ is the entropy function. We add a regularization term to our network and the clustering objective is $\ell_{C}$:
\begin{equation}
    \ell_{C}= \frac{1}{N}\sum_{i=1}^{N} h(\sigma(f_{\theta}(x_i))) - h(\frac{1}{N} \sum_{i=1}^{N} \sigma(f_{\theta}(x_i))) + \alpha \sum_{l=1}^L {\Vert{{\theta}^l}\Vert{}}^2
\label{eq: mutual_information}
\end{equation}
\vskip -0.1in
where $\alpha$ denotes the hyper-parameter for network parameters $\{\theta^{1} \ldots \theta^{L}\}$. Maximizing $H(Y)$ will punish imbalanced cluster size and prevent trivial solutions where all the instances are clustered into one cluster whilst minimizing $H(Y|X)$ will map similar instances $x$ to have similar labels $y$. Note we favor this probabilistic discriminative clustering model since it has fewer assumptions about the natures of categories are made \cite{krause2010discriminative} and fits our fairness objective which requires fractional clustering assignments as inputs to indicate the
degree of cluster assignment belief.

\subsection{Generating Fair Assignments Under Group-level Fairness Constraints}
Let the fractional clustering assignments from the current learned model be $Y = \{ y_1, \ldots y_N\} \in {\R}^{N \times K}$. To use these to form fair clustering assignments, we solve a fairer assignment matrix $\hat{Y} = \{\hat{y}_1, \ldots \hat{y}_N \}\in {\Z}^{N \times K}$ that satisfy our optimal fairness condition: $P(x \in G_t | x \in C_k) = \rho_t$. 
To address the fair assignment problem we formulate our fairer assignment problem as an integer linear programming problem where we aim to minimize the changes to the current assignment $Y$ to obtain a fairer assignment $\hat{Y}$ as follows:
\begin{equation}
\textbf{Objective:}~~\\
 \argmin_{\hat{Y}} \sum_{i=1}^{N} [1 - y_i * {\hat{y_i}}^T] 
 \label{p1}
\end{equation}
Recall that $y_i$ is a probability distribution over the cluster assignments for instance $i$ and $\hat{y_i}$ chooses exactly one cluster to assign instance $i$ to. Naturally the objective is maximized when $y_i$ is assigned to its most probable cluster but this may cause an unfair clustering.

We denote $\rho_i = {|G_i|}/{N}$ as the fraction of the protected group $G_i$ in the data set and our aim is for each cluster to have the same density. Let $M \in {\Z}^{N \times T}$ encode the sensitive attributes for the entire population such that $M_{it} \in \{0, 1\}$ indicates whether an instance $x_i$ belongs to a protected group $G_t$. To satisfy optimal fair condition $P(x \in G_t | x \in C_k) = \rho_t$ we have the following constraints:
\begin{equation}
 \sum_{i=1}^{N} M_{it} \hat{y}_{ij} = \sum_{i=1}^{N}\hat{y}_{ij} \rho_t \quad \forall j \in \{1 \ldots K\},t \in \{1 \ldots T\}
 \label{p2}
\end{equation}

Now we relax the problem by fixing the size of each new cluster to make the constraint matrix totally unimodular. 
Let the rounded version of current assignment $Y$ as $Y^{'}$ then the size of cluster $C_j$ is $|C_j| = \sum_{i=1}^{N} y_{ij}^{'}$. The constraints for new clusters' size are:
\vskip -0.1in
\begin{equation}
    \sum_{i=1}^{N}\hat{y}_{ij} = |C_j|  \quad \forall j \in \{1 \ldots K\} 
 \label{p3}
\end{equation}
\vskip -0.1in
Lastly we add constraints for $\hat{Y}$ to ensure each instance is assigned to one cluster:
\begin{equation}
\sum_j \hat{y}_{ij} = 1 \quad \forall i \in \{1 \ldots N\}
 \label{p4}
\end{equation}
\vskip -0.1in
Note this ILP formulation also supports user-defined $\rho_t$ which can be seen as a flexible fairness rule.  
Next we show the constraint matrix of our ILP problem is totally unimodular so that we can efficiently solve it with a LP solver and still return integral solutions.

\paragraph{Proof of Total Unimodularity of Constraint Matrix:} We know that if a constraint matrix of an ILP is totally unimodular (TU) then we can solve the problem using an LP (linear program) solver and the solution will still be integral \cite{schrijver1998theory}. Using an LP solver will largely reduce the running time and \cite{vaidya1989speeding} has shown that the running time for LP is polynomial in the input size.  
 
In the above proposed constraints, there are $NK$ unique regular variables ($N$ instances and $K$ categories). To construct the constraint matrix $C$ which encodes constraint \ref{p2}, \ref{p3} and \ref{p4}, we will use $2NK$ regular variables. Matrix $C$ has $T + 1$ rows (the first $T$ rows correspond to the fairness constraints in equation \ref{p2} and last row corresponds to constraints in equation \ref{p4}) and $N+K$ columns. Note the first $K$ columns of the last row are set to $0$ and the last $N$ columns of first $T$ rows are set to $0$. In matrix $C$, each entry of $C$ is from $\{-1, 0, 1\}$. Moreover, each column only has one non-zero element. This is because: (1) for constraints set in equation \ref{p2}, each instance only belongs to one protected group, (2) for constraints set in equation \ref{p4}, there is only one row vector with $K$ elements as $1$ to ensure the valid assignment.  

\begin{lemma} \textbf{TU Identity} \cite{schrijver1998theory}.
Let $C$ be a matrix such that all its entries are from $\{0,1,-1\}$. Then $C$ is totally unimodular, i.e., each square submatrix of $C$ has determinant $0$, $1$, or $-1$ if every subset of rows of $C$ can be split into two parts $A$ and $B$ so that the sum of the rows in $A$ minus the sum of the rows in $B$ produces a vector all of whose entries are from $\{0,1,-1\}$.
\label{lemma2}
\end{lemma}
\begin{theorem}
\label{thm:eff}
The matrix $C$ formed by the coefficients of the constraints used to encode our proposed constraints from equation \ref{p2}, \ref{p3} and equation \ref{p4} is totally unimodular.
\end{theorem}
\renewcommand{\algorithmicrequire}{\textbf{Input:}}
\renewcommand{\algorithmicensure}{\textbf{Output:}}
\begin{algorithm}[t]
\small
 \caption{Main learning algorithm for deep fair discriminative clustering.}
 \label{alg:1}
\begin{algorithmic}[1]
\REQUIRE Input $\{x_k\}_{k=1}^N$, sensitive attributes $M$, cluster size $K$, network structure $f$, hyper-parameters $\alpha,\beta,\gamma$.
\ENSURE Clustering network $f_{\theta}$, predictions $\{y_k\}_{k=1}^N$ .
 \FOR{each pre-trained epoch}
 \FOR{sampled mini-batch $\{x_k\}_{k=1}^n$}
 \STATE Calculate $\ell_{C}= \frac{1}{n}\sum_{i=1}^{n} h(\sigma(f_{\theta}(x_i))) - 
 $ $h(\frac{1}{n} \sum_{i=1}^{n} \sigma(f_{\theta}(x_i))) + \alpha \sum_{l=1}^L {\Vert{{\theta}^l}\Vert{}}^2$
 \STATE Generate $x_{k}^{'} = x_{k} + t$ via solving $t$ from eq \ref{eq:vat}.
 \STATE Calculate $\ell_{Aug} = \sum_{i=1}^n \text{KL} (\sigma(f_{\theta}(x_i)) , \sigma(f_{\theta}({x_i^{'}})))$
  \STATE Update network $f_{\theta}$ via minimizing $\ell_C + \gamma \ell_{Aug}$.
 \ENDFOR
 \ENDFOR
 \REPEAT
 \STATE Generate predictions $\{y_k\}_{k=1}^N$ based on $f_{\theta}$.
 \STATE Construct a fair assignment problem via objective \ref{p1} and constraints defined in eq \ref{p2}, \ref{p3} and \ref{p4}.
 \STATE Solve fair assignments $\{\hat{y}_k\}_{k=1}^N$ via LP solver.
 \FOR{sampled mini-batch $\{x_k\}_{k=1}^n$}
 \STATE Calculate $\ell_{Fair} = \frac{1}{n}\sum_{i=1}^n \hat{y}_{i}log(\sigma(f_{\theta}(x_i)))$
 \STATE Calculate $\ell_{C}= \frac{1}{n}\sum_{i=1}^{n} h(\sigma(f_{\theta}(x_i))) - 
 $ $h(\frac{1}{n} \sum_{i=1}^{n} \sigma(f_{\theta}(x_i))) + \alpha \sum_{l=1}^L {\Vert{{\theta}^l}\Vert{}}^2$
 \STATE Generate $x_{k}^{'} = x_{k} + t$ via solving $t$ from eq \ref{eq:vat}.
 \STATE Calculate $\ell_{Aug} = \sum_{i=1}^n \text{KL} (\sigma(f_{\theta}(x_i)) , \sigma(f_{\theta}({x_i^{'}})))$
 \STATE Calculate $\ell =  \ell_{C} + \beta \ell_{Fair} + \gamma \ell_{Aug}$
 \STATE Update network $f_{\theta}$ via minimizing $\ell$.
 \ENDFOR
\UNTIL{$\{y_k\}_{k=1}^N$ satisfy \emph{optimal fairness} rules}
\end{algorithmic}
\end{algorithm}
\subsection{Learning to Be Fairer with Contrastive Learning}
To learn a fair clustering model we aim to exploit the fairness assignments $\hat{Y}$ to reshape the features learned via clustering networks $f_{\theta}$.  We treat $\hat{Y}$ as “pseudo-labels” to optimize the following cross entropy loss $\ell_{Fair}$ for fairer results:
\begin{equation}
\ell_{Fair} = \frac{1}{N} \sum_{i=1}^{N} \sum_{j=1}^{K} \hat{y}_{ij}log y_{ij} = \frac{1}{N}  \sum_{i=1}^{N}\hat{y}_{i}log(\sigma(f_{\theta}(x_i)))
    \label{eq:fairness}
\end{equation}
\vskip -0.1in
Simply optimizing the fairness loss $\ell_{Fair}$ will dramatically change the current clustering representations to fit an approximated fair assignment $\hat{Y}$ which harms the clustering properties. Instead, we propose to learn a fairer and clustering-friendly representation simultaneously by combining the clustering loss $\ell_C$ with fairness loss $\ell_{Fair}$. Note the fair assignments $\hat{Y}$ are updated after each training epoch as the ``nearest" fair assignments for current clustering predictions. 

Inspired by the recent use of contrastive learning \cite{he2020momentum,chen2020simple}  which leads to state-of-the-art performance in the unsupervised training. Here we consider a general augmentation strategy beyond classic augmentation techniques within vision domain: we propose to add a small local perturbation of instance $x$ such that $x^{'} = x + t$ and hope to maximize the perturbation $t$ subject to the constraint that the clustering assignments for $x$ and $x^{'}$ are the same. To achieve this goal we apply the virtual adversarial training \cite{miyato2018virtual} work to generate adversarial direction for perturbation $t$. Denote the current model's parameters $\theta$ to help estimate the true clustering indicator vector for instance $x$ as $\sigma({f_{{\theta}}(x)})$, the formulation to compute the adversarial perturbation $t_{adv}$ is as follows:
\begin{equation}
    t_{adv} = \argmax_{t; {||t||}_2 \le \epsilon} \text{ KL}(\sigma{(f_{{\theta}}(x))}, \sigma{(f_{\theta}(x + t))})
    \label{eq:vat}
\end{equation}
Note that $\epsilon$ is a scalar-valued hyper-parameter with constraint that $\epsilon > 0$. \cite{miyato2018virtual} has shown that solving $t_{adv}$ can be approximated efficiently via few gradient steps. With the generated $t_{adv}$ we propose the augmentation loss $\ell_{Aug}$ which minimizes the Kullback–Leibler divergence between clustering assignment $\sigma(f_{\theta}(x_i))$ and its augmented version's assignment $\sigma(f_{\theta}({x_i}^{'}))$:
\begin{equation}
        \ell_{Aug} = \sum_{i=1}^N \text{KL} (\sigma(f_{\theta}(x_i)) , \sigma(f_{\theta}({x_i}^{'})))
\label{eq:SAT}
\end{equation}
\vskip -0.1in
To train our proposed framework, we start with the training on clustering network $f(\theta)$ via optimizing the clustering loss $\ell_{C}$ and augmentation loss $\ell_{Aug}$ to ensure the data are separated into different meaningful clusters; as clustering model converges we generate fair assignments after each training epoch based on the objective in eq \ref{p1} and optimize the overall loss function $\ell$ by concatenating the fairness loss $\ell_{Fair}$ to augmentation and clustering objectives as $\ell =  \ell_{C} + \beta \ell_{Fair} + \gamma \ell_{Aug}$ where $\beta, \gamma$ are positive weight hyper-parameters. Algorithm \ref{alg:1} summarizes the proposed method.


\section{Experiments}
In this section, we conduct experiments\footnote{In our experimental work we use the fair clustering data sets used by earlier work for comparison.} to evaluate our approach empirically. Based on our experiments, we report the following key results:
\begin{itemize}
 \item In typical fair clustering settings our approach achieves better clustering performance and near-optimal fairness results compared against existing baselines. 
 \item Our approach is effective in novel fair clustering settings such as flexible fairness constraints, clustering with multi-state PSVs, and predictive clustering.
\item We show our learned embedding converges to a space useful for fairness clustering quickly and also provides insights on tuning hyper-parameter $\beta$ to achieve our fairness goal with a minimum loss on clustering performance.
\end{itemize}

\subsection{Experimental Setup}
\noindent
\paragraph{Deep Clustering Data Sets.} We first evaluate our work on two visual data sets with binary PSV that has been used in recent work \cite{li2020deep}: 1) MNIST-USPS consists of $67291$ training images of hand-written digits. We use the  image source (MNIST or USPS) as a binary PSV and cluster the data into $10$ classes representing $10$ digits. 2) Reverse-MNIST takes the $60000$ training images from MNIST and creates an inverted duplicate to build this dataset. The binary PSV is then original or inverted and the total number of classes is $10$. Moreover, we evaluate one challenging fair clustering task with multi-state PSV on the HAR dataset used in \cite{wang2019towards}: 3) HAR 
contains $10299$ instances in total with captured action features for $30$ participants. There are $6$ actions in total which serve as labels for clustering. The identity of each person is used as the PSV value. 

\noindent
\paragraph{Classic Data Sets.}
Following traditional fair clustering work \cite{bera2019fair} we choose three tabular datasets for complete comparison: 1) Census 
data with $5$
attributes (``age'', ``fnlwgt'', `education-num'', ``capitalgain'', ``hours-per-week'') and binary PSV gender, we set whether income exceeds $50$K as the clustering label; 2) Bank 
data with $3$ attributes (``age'', ``balance'', ``duration-of-account'') and binary PSV marital, we set whether a client will subscribe a term deposit as the label; 3) Credit 
data with $14$ features and binary PSV marital, we set whether the cardholder will make a payment as the label.

\paragraph{Measurements.}
To measure the clustering quality for deep fair clustering and other baselines, we use both clustering accuracy (Acc) \cite{xu2003document,yang2010image} and normalized mutual information (NMI) metrics for a comprehensive study. To evaluate the fairness, we use the balance measure defined in equation \ref{eq:old_fair_multi}. For all those three measures, higher values indicate better performance. 

\paragraph{Algorithms.}
For the deep clustering baselines, we use DEC \cite{xie2016unsupervised} as a representative method for centroid-based clustering and IMSAT \cite{hu2017learning} for discriminative clustering approach. For fair clustering algorithms, we choose the scalable fair clustering algorithm \cite{backurs2019scalable} and the fair algorithms for clustering \cite{bera2019fair}. For deep fair clustering baselines, we compare our work with the latest work \cite{li2020deep} and the geometric-based fair clustering \cite{wang2019towards}. For our own approach, we set parameters $\alpha =10^{-4}$, $\gamma = 1$ and $\beta = 4$. For the structure of $f_{\theta}$, we use two convolutional layers followed by batch normalization and pooling for visual data and fully connected layers for non-visual data. 
For a fair comparison with non-deep clustering baselines, we use pre-trained auto-encoder's features like \cite{li2020deep}. For the ILP objective we use the Gurobi \cite{gurobi} optimizer. More details are included in the \href{https://deep-fair-descriptive-clustering.s3-us-west-2.amazonaws.com/Supplementary-Material-for-DFDC.pdf}{\emph{supplementary material}}.

\subsection{Evaluation}
 \begin{table*}[ht]
 \small
  \centering
    \caption{Comparison of clustering and fairness performance on MNIST-USPS, Reverse-MNIST and HAR. HAR consists of multi-state PSV that baselines with dashes are not applicable. Bold results are the best results among all the baselines except the ground-truth and the guaranteed fairness results which are marked with blue. Note we report our average performance results after $10$ trials and the term \emph{optimal} refers to the clustering giving the optimal accuracy and the corresponding balance.}
  \scalebox{0.81}{
  \begin{tabular}{p{1.78cm}|p{3.38cm}|ccc|ccc|ccc}
  \hline
    \multicolumn{2}{c|}{}&\multicolumn{3}{|c|}{\textbf{MNIST-USPS}}&\multicolumn{3}{|c|}{\textbf{Reverse-MNIST}}&\multicolumn{3}{|c}{\textbf{HAR}} \\
    \hline
     \textbf{Category}&\textbf{Methods}&Acc & NMI & Balance &Acc &  {NMI} & {Balance}&{Acc} &  {NMI} & {Balance}\\
    \hline
    &Ground Truth (Optimal)& 1.000 & 1.000  &\textcolor{blue}{0.120}  &1.000 &1.000 &\textcolor{blue}{1.000} &1.000   &1.000   &{0.458}\\
    \hline
    \multirow{2}{1.9cm}{Non-Fair Deep Clustering}&DEC     \cite{xie2016unsupervised}     & 0.586 &0.686  &0.000  &0.401  &0.480  &0.000  &0.571  &0.662  &0.000\\
    &IMSAT   \cite{hu2017learning}   & 0.804 &0.787  &0.000  &0.525  &0.630  &0.000  &0.812  &0.803  &0.000\\
    \hline
    \multirow{2}{1.9cm}{Fair Non-Deep Clustering}&ScFC    \cite{backurs2019scalable}   & 0.176 &0.053  &\textcolor{blue}{0.120}  &0.268  &0.105  &\textcolor{blue}{1.000}  & --  &--  &--\\
    &FAlg    \cite{bera2019fair}& 0.621 &0.496  &0.093  &0.295  &0.206  &0.667  &0.642  &0.618  &0.420\\
    \hline
         \multirow{4}{1.9cm}{Fair Deep  Clustering}&\cite{wang2019towards}& 0.725 &0.716  &0.039  &0.425  &0.506  &0.430  &0.607  &0.661  &0.166\\
   &DFCV    \cite{li2020deep}& 0.825 &0.789  &0.067  &0.577  &0.679  &0.763  &--  &--  &--\\
    &Initial ILP Results (Ours)   & 0.779 &0.746  &\textcolor{blue}{0.120}  &0.485  &0.577  &\textcolor{blue}{1.000}  &0.789  &0.702  &\textcolor{blue}{0.653}\\
    &\textbf{Ours Result}    & \textbf{0.936} &\textbf{0.876}  &\textbf{0.119}  &\textbf{0.589}  &\textbf{0.690}  &\textbf{0.946}  &\textbf{0.862}  &\textbf{0.845}  &\textbf{0.468}\\
    \hline
\end{tabular}}
    \label{tab:performance}
\end{table*}
\paragraph{High Dimensional Data.} As shown in the Table \ref{tab:performance}, traditional fair clustering algorithms achieve good fairness results especially  ScFC which returns guaranteed fair clusters. However the clustering performance is not good as deep clustering methods due to the lack of representation learning 
Both DEC and IMSAT achieve reasonable clustering results but poor fairness results, this shows the unfairness of existing deep clustering methods which motivates our adding fairness rules. Comparing our results with the recent deep fair clustering works \cite{wang2019towards,li2020deep} we can see that our approach consistently outperforms these two baselines in terms of both clustering performance and fairness. Observing the ground truth results in Table \ref{tab:performance} we can see the fairness rules can be seen as \emph{positive guidance} to improve the clustering performance, our approach is shown to be able to learn from this guidance and improve fairness as well as accuracy. Comparing our final results with initial fair ILP assignments, we demonstrate the success of iteratively learning fairer and clustering-favored features.

 \begin{table}[t]
  \centering
  \caption{Measuring our work's ability to obtain optimum fairness on Bank, Census and Credit data sets. Ours w/o fairness denotes the results from our proposed model without the fairness objective.}
  \scalebox{0.9}{
  \begin{tabular}{c|c|c|c|c|c|c|c}
  \hline
    &\multicolumn{2}{|c|}{\textbf{k-Means}}&\multicolumn{2}{|c|}{\textbf{Ours w/o fairness}}&\multicolumn{2}{|c|}{\textbf{Ours with fairness}}&\textbf{Optimal} \\
    \hline
     &{Acc} & {Balance} &{Acc} & {Balance}&{Acc}& {Balance}&Balance \\
    \hline
    Census   & \text{0.647} & \text{0.406}  &\text{0.649}  &\text{0.401}  &\text{0.620}  &\text{0.489}  &\text{0.492}\\
    \hline
    Bank     & \text{0.656} &\text{0.144}  &\text{0.664}  &\text{0.144}  &\text{0.645}  &\text{0.186}  &\text{0.189}\\
    \hline
    Credit   & \text{0.694} &\text{0.835}  &\text{0.697}  &\text{0.828}  &\text{0.690}  &\text{0.855}  &\text{0.856} \\
    \hline
\end{tabular}}
\label{tab:tabular_data}
\end{table}
\paragraph{Classic Tabular Datasets.}
 We evaluate our approach on tabular data sets used by classic fair clustering algorithms and present the results in Table \ref{tab:tabular_data}. We find that our model achieves similar results as k-Means on tabular data with human-defined semantic features. Meanwhile, both k-Means and our deep clustering model achieve good balance results when the number of clusters is correctly set as the ground truth number of classes. Even when the initial balance is high, our model can still largely improve it and achieve near-optimal balance with a slight loss in terms of clustering accuracy.


\paragraph{Predictive Clustering Results.} Here we evaluate our method's ability to make predictions on test data without PSV information which is a new setting in the fair clustering literature. That is we have already clustered a data set with PSV values and are now making predictions using the model learnt. This is particularly important for practitioners who are, for instance, deploying models on the web (where individuals are reluctant to share PSV information) and we see our results in Table \ref{tab:predictive_data}. Our approach performs well in both train and test sets. One exception is that the test balance in HAR is much lower than the training balance, we hypothesize this is due to different distributions between training and test set, the optimal test balance is $1$ while the optimal training balance is $0.653$. 

 \begin{table}[t]
  \centering
   \vskip -0.1in
    \caption{Our clustering results on the test sets of high-dimensional data without the sensitive attributes information. The optimal balance denotes the test set's optimal results.}
  \scalebox{0.9}{
  \begin{tabular}{c|c|c|c|c|c|c|c}
  \hline
    &\multicolumn{2}{|c|}{\textbf{Acc}}&\multicolumn{2}{|c|}{\textbf{NMI}}&\multicolumn{3}{|c}{\textbf{Balance}} \\
    \hline
     &{train} & {test} &{train} & {test}&{train}& {test}&{optimal} \\
    \hline
    MNIST-USPS   & \text{0.936} & \text{0.930}  &\text{0.876}  &\text{0.865}  &\text{0.119}  &\text{0.164} &\text{0.200}\\
    \hline
    Reverse-MNIST     & \text{0.589} &\text{0.562}  &\text{0.690}  &\text{0.645}  &\text{0.946}  &\text{0.890}&\text{1.000} \\
    \hline
    HAR   & \text{0.862} &\text{0.831}  &\text{0.845}  &\text{0.795}  &\text{0.468}  &\text{0.167}&\text{1.000}  \\
    \hline
\end{tabular}}
\vskip -0.15in
    \label{tab:predictive_data}
\end{table}

\subsection{Further Analysis on Our Model}
Here we explore  our model's flexibility in satisfying fairness requirements and better understand its performance by feature space visualization and empirical convergence.
 \begin{table}[h]
 \vskip -0.2in
  \centering
    \caption{Results on flexible fair constraints for MNIST-USPS data.}
  \scalebox{0.9}{
  \begin{tabular}{c|c|c|c}
  \hline
    \textbf{Relax Degree} &\textbf{Acc} & \textbf{NMI} & \textbf{Balance}\\
    \hline
    $\epsilon = 0.01$     & \text{0.903} & \text{0.847}  & \text{0.103} \\
    \hline
    $\epsilon = 0.02$&  \text{0.892} & \text{0.835}  & \text{0.095} \\
    \hline
     $\epsilon = 0.03$   &  \text{0.869} & \text{0.818}  & \text{0.081}\\
    \hline
    $\epsilon = 0.04$   &  \text{0.847} & \text{0.798}  & \text{0.069}\\
    \hline
\end{tabular}}
\vskip -0.2in
\label{tab:novel}
\end{table}

\paragraph{Results on Flexible Fairness Constraints.}
Here we explore how relaxing the optimal fair condition defined as $\rho_{t}$ in equation \ref{p2} produces flexible constraints. We now require the fairness requirement to be in the interval $[\rho_{t} - \epsilon, \rho_{t} + \epsilon]$. In Table \ref{tab:novel}, we can see a larger relaxation $\epsilon$ leads to a lower balance which is expected; since the fairness signals can serve as positive guidance for clustering in MNIST-USPS, we observe the Acc and NMI are decreasing with larger relaxation. Allowing flexible constraints are important from a practitioner's perspective as the fairness rules vary across regions. 

 \begin{figure*}[t]
 \centering
 \subfloat[epoch 0]{
 \includegraphics[width=0.18\textwidth]{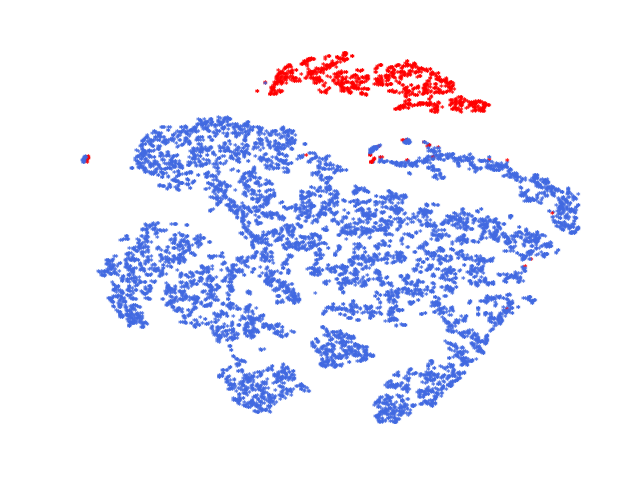}}
 \label{4a}
 \hfill
 \subfloat[epoch 15]{
 \includegraphics[width=0.18\textwidth]{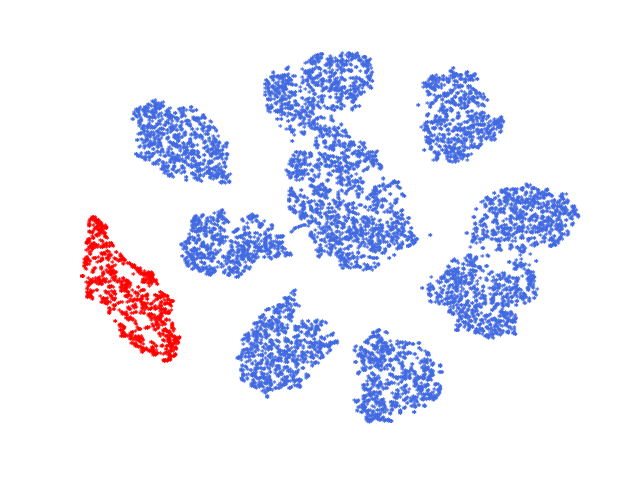}}
 \label{4b}
 \hfill
  \subfloat[epoch 30]{
 \includegraphics[width=0.18\textwidth]{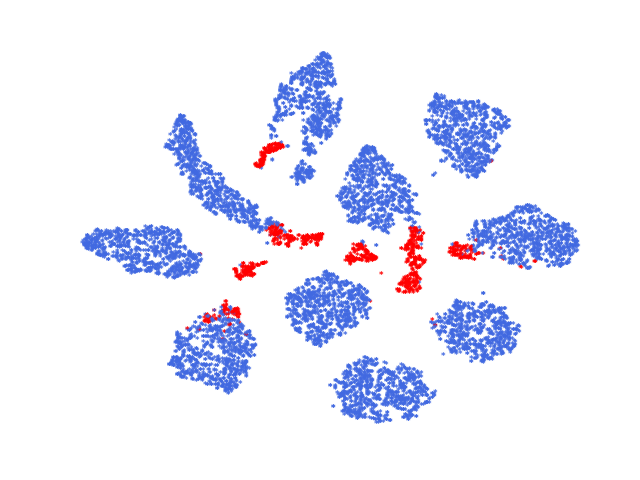}}
 \label{4c}
 \hfill
  \subfloat[epoch 45]{
 \includegraphics[width=0.18\textwidth]{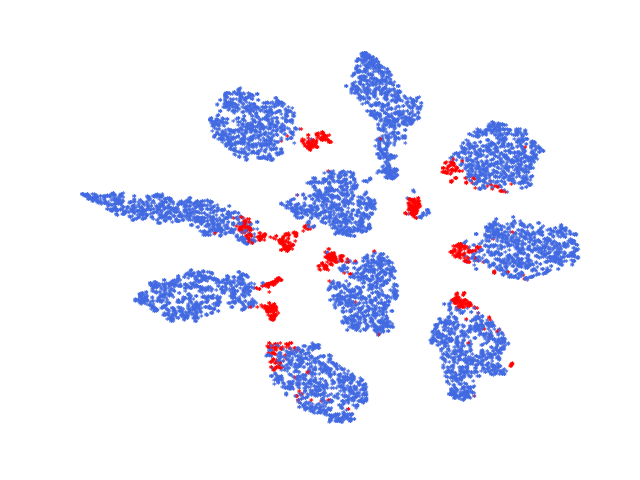}}
 \label{4d}
 \hfill
  \subfloat[epoch 60]{
 \includegraphics[width=0.18\textwidth]{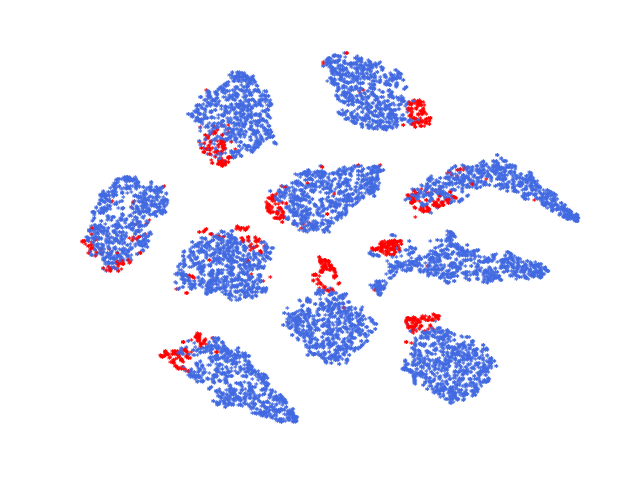}}
 \label{4e}
\caption{t-SNE visualization of learned embedding, color red and blue indicate different PSV values. }
\label{fig:visualization}
\end{figure*}

 \begin{figure}[t]
 \centering
 \subfloat[MNIST-USPS (Balance)]{
 \includegraphics[width=0.24\textwidth]{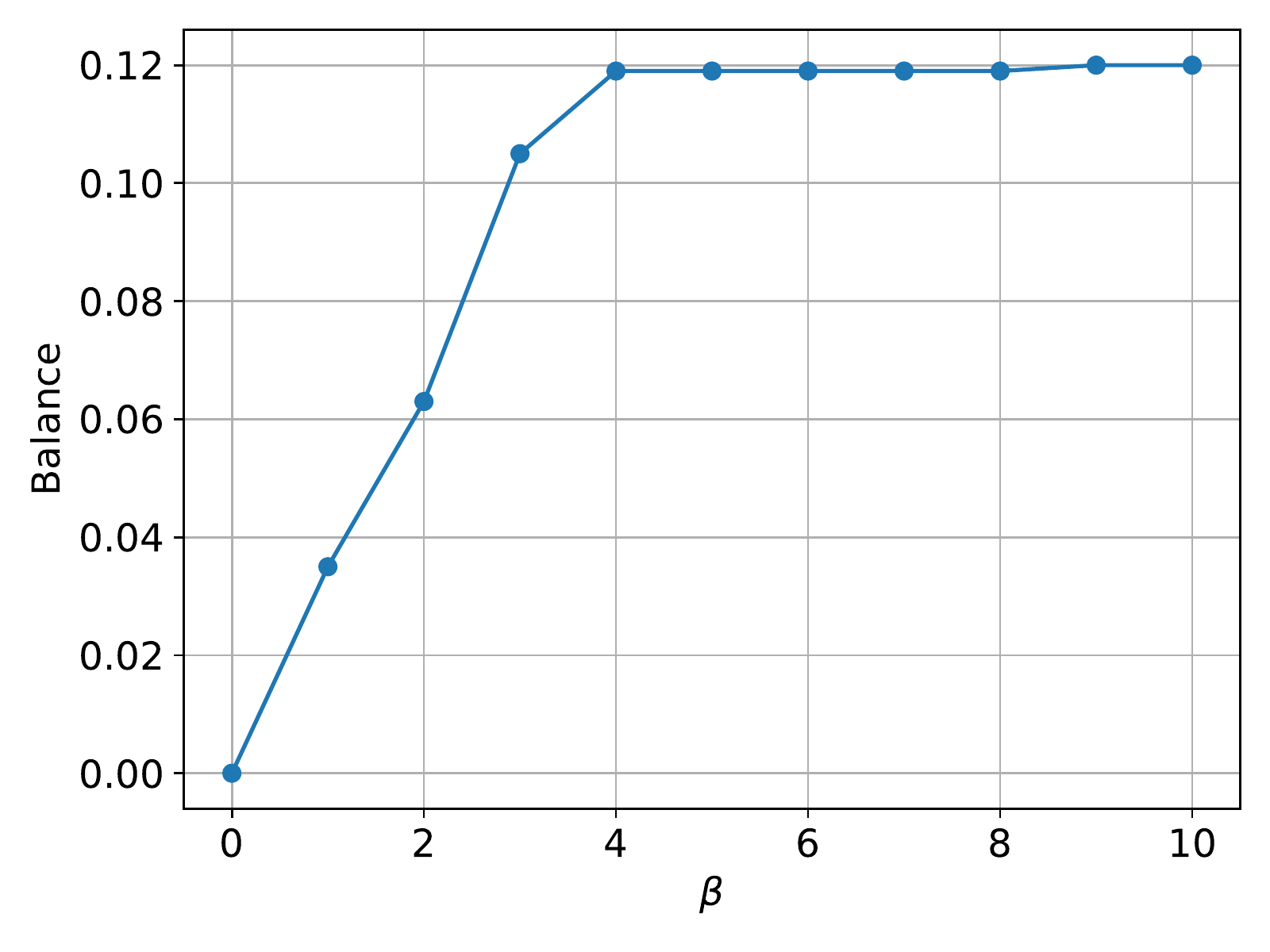}}
 \label{3aa}
 \hfill
 \subfloat[MNIST-USPS (Acc)]{
 \includegraphics[width=0.24\textwidth]{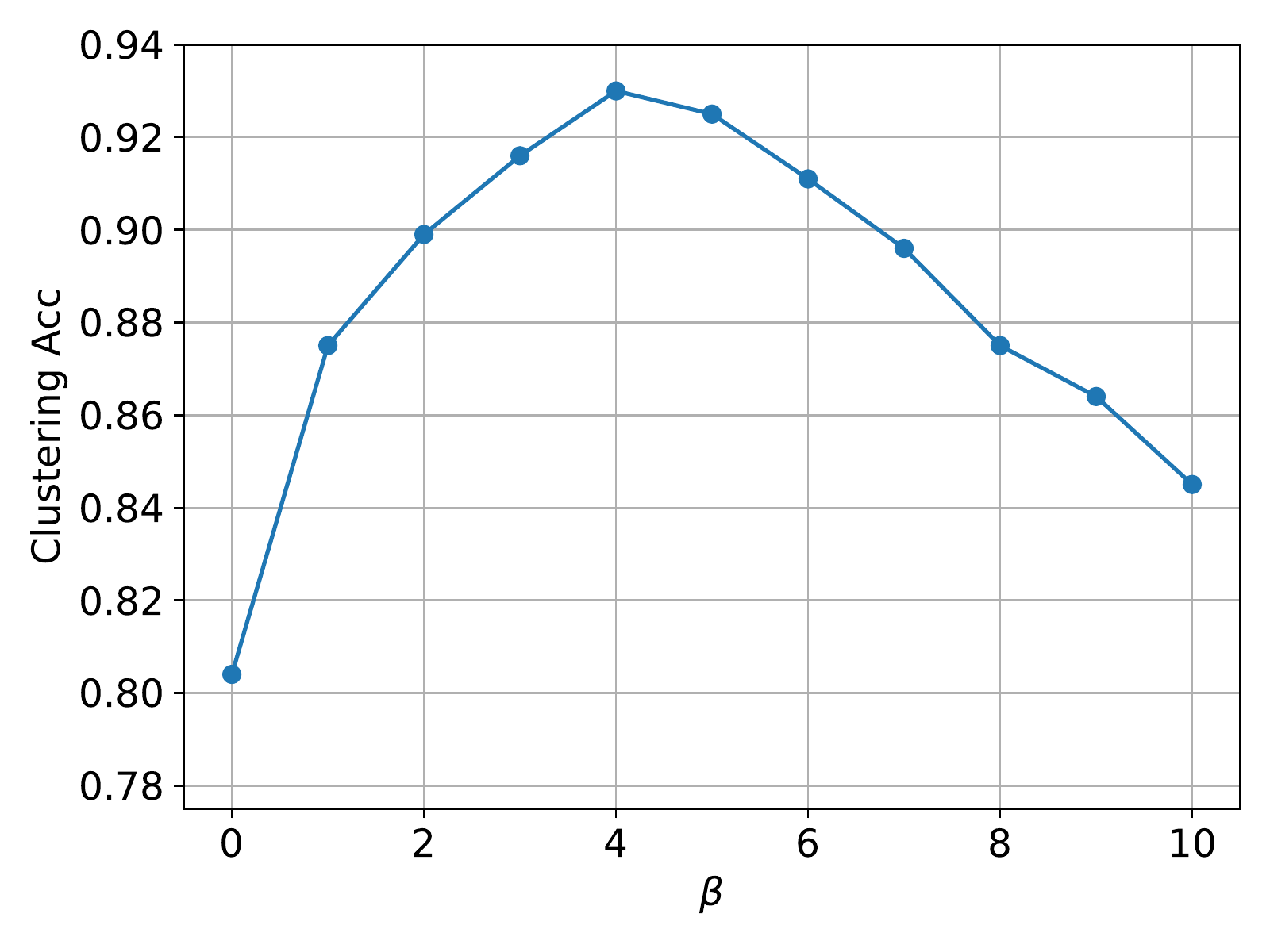}}
 \label{3bb}
  \subfloat[HAR (Balance)]{
 \includegraphics[width=0.235\textwidth]{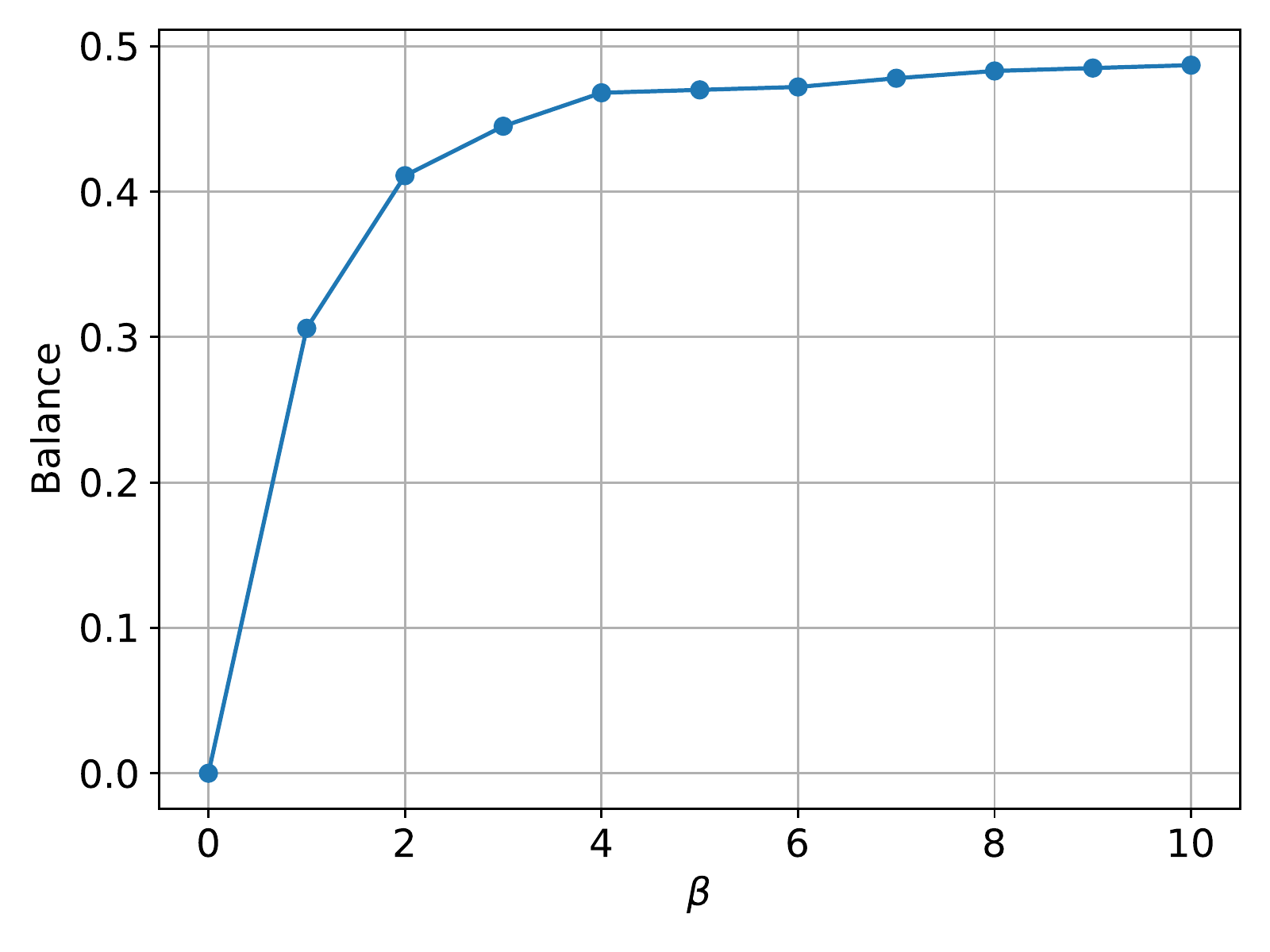}}
 \label{3cc}
 \hfill
 \subfloat[HAR (Acc)]{
 \includegraphics[width=0.24\textwidth]{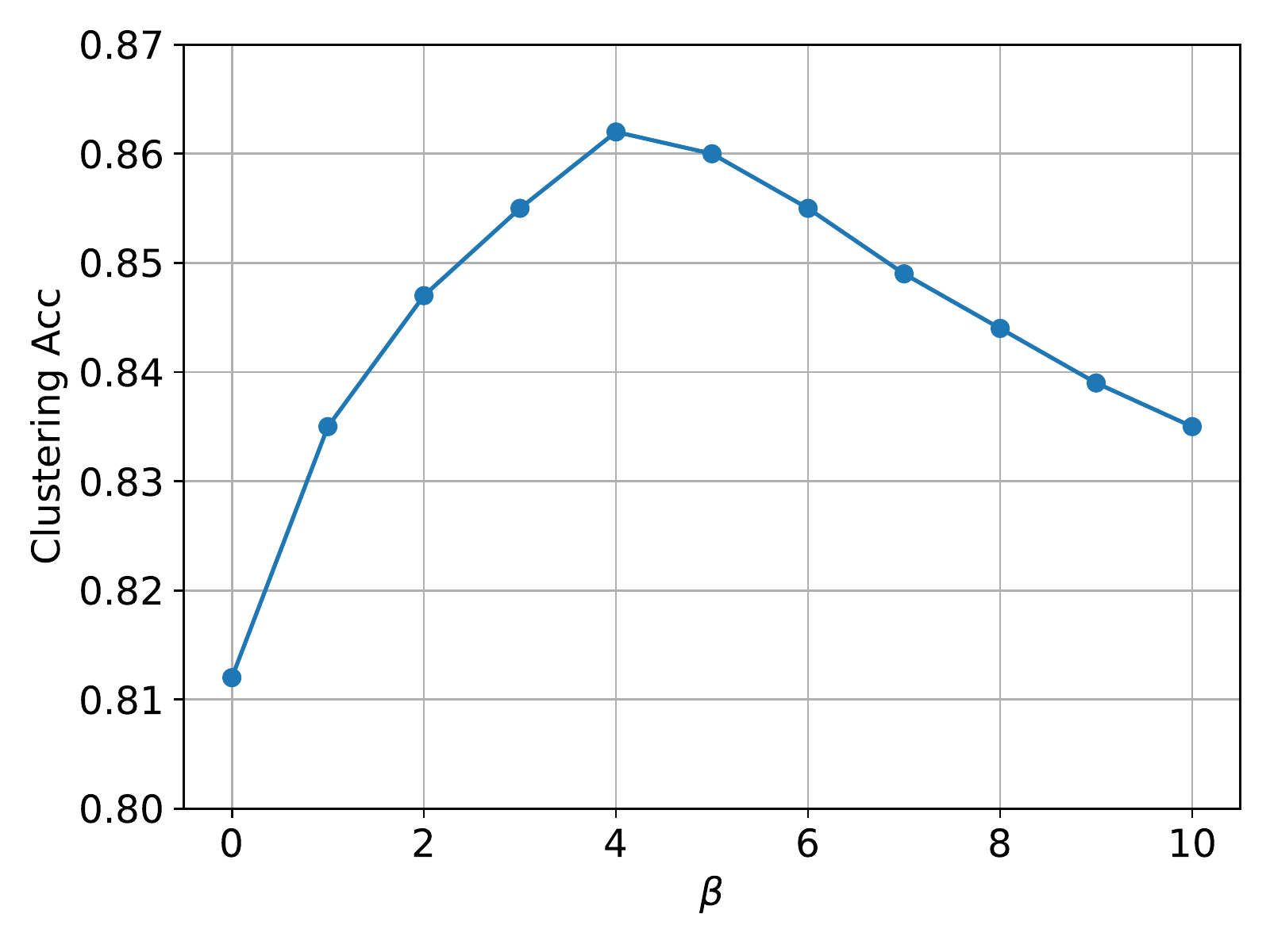}}
 \label{3dd}
\caption{Sensitivity analysis of fairness hyper-parameter $\beta$.}
 \vskip -0.1in
\label{fig:parameter}
\end{figure}

\textbf{Feature Space Visualizing.}
To understand how our model learns a fair representation, we have applied t-SNE \cite{van2008visualizing} to visualize the feature space of MNIST-USPS during different training epochs in Figure \ref{fig:visualization}. The initial model is trained with deep clustering objectives which yield unfair results, once we introduce fairness objectives the red instances start to move to different clusters. Meanwhile we can find our learned representations still maintain good clustering performance. 

\textbf{Tuning the Weight of Fairness Objective.}
We experiment on the choices of hyper-parameter $\beta$ which controls the weight of the fairness objective and report the clustering results in Figure \ref{fig:parameter}. It is straightforward to see from (a) and (c) that as $\beta$ increases, the training balance increases. Meanwhile, based on (b) and (d) we can find the Acc goes up and down as $\beta$ increases.
Our previous result shows that the fairness constraints can serve as positive guidance for both MNIST-USPS and HAR. That is why the clustering accuracy goes up when we increase $\beta$ from 0. But we also observe that with a very large $\beta$ the clustering accuracy will drop. We hypothesize this is because the fairness objective dominates the overall objective so that the impact of clustering objective is hindered. As balance can be tracked during the training process for free, our insight for selecting hyper-parameter $\beta$ is to pick the smallest $\beta$ that achieves satisfying balance results.

\begin{minipage}{0.49\textwidth}
\textbf{Empirical Convergence Analysis.} To investigate the smoothness of learning with clustering and fairness objectives together, we present the learning curves of overall training loss and the balance results in Figure \ref{fig:convergence_loss} and \ref{fig:convergence_balance}. We can see from the plots that the train loss drops quickly and converges after $50$ epochs. Our model's balance result also converges after $50$ epochs.  
\end{minipage}
\begin{minipage}{0.25\textwidth}
 \includegraphics[width=\textwidth]{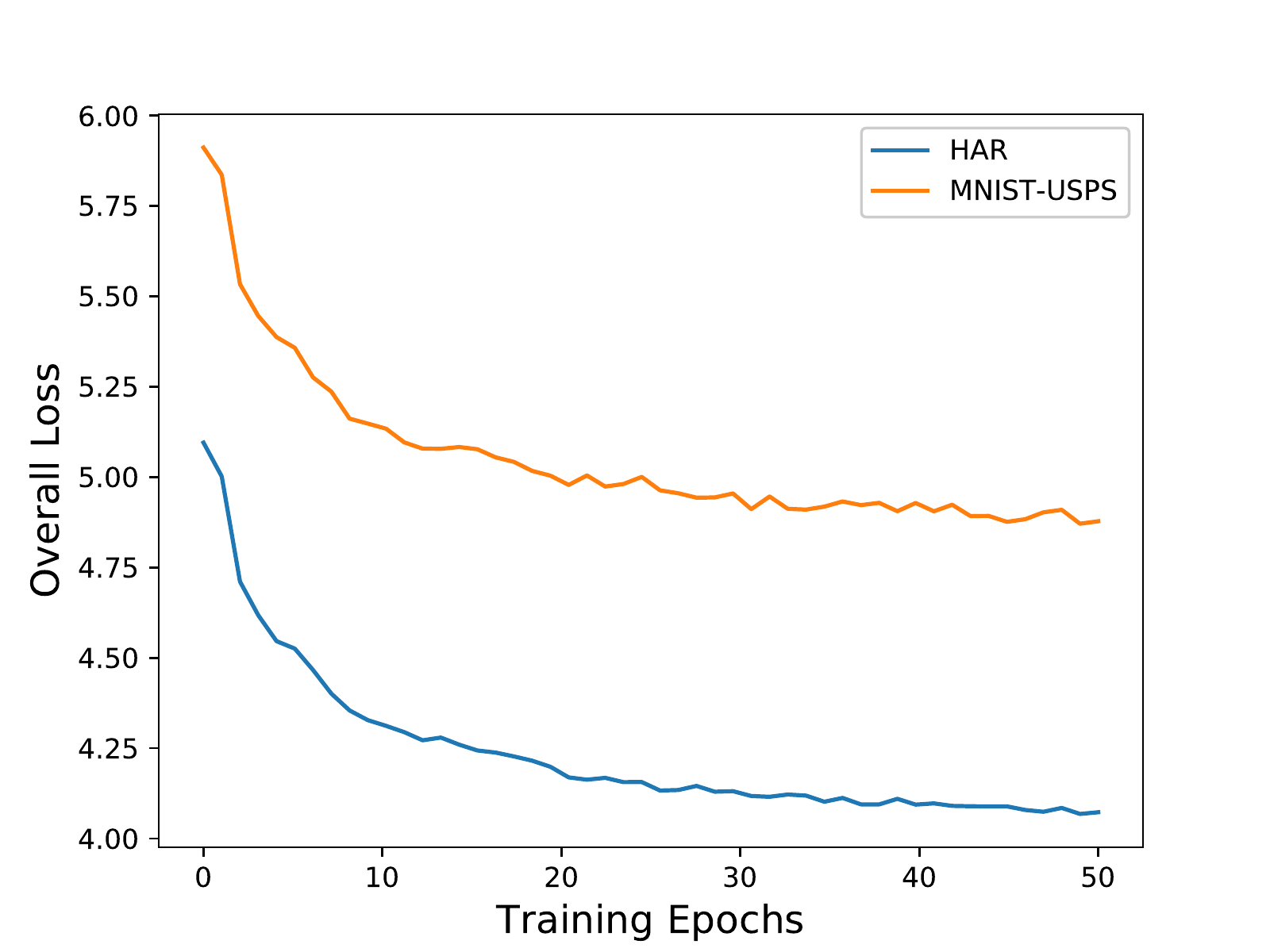}
 \captionof{figure}{Train Loss}
 \label{fig:convergence_loss}
\end{minipage}
\begin{minipage}{0.25\textwidth}
 \includegraphics[width=\textwidth]{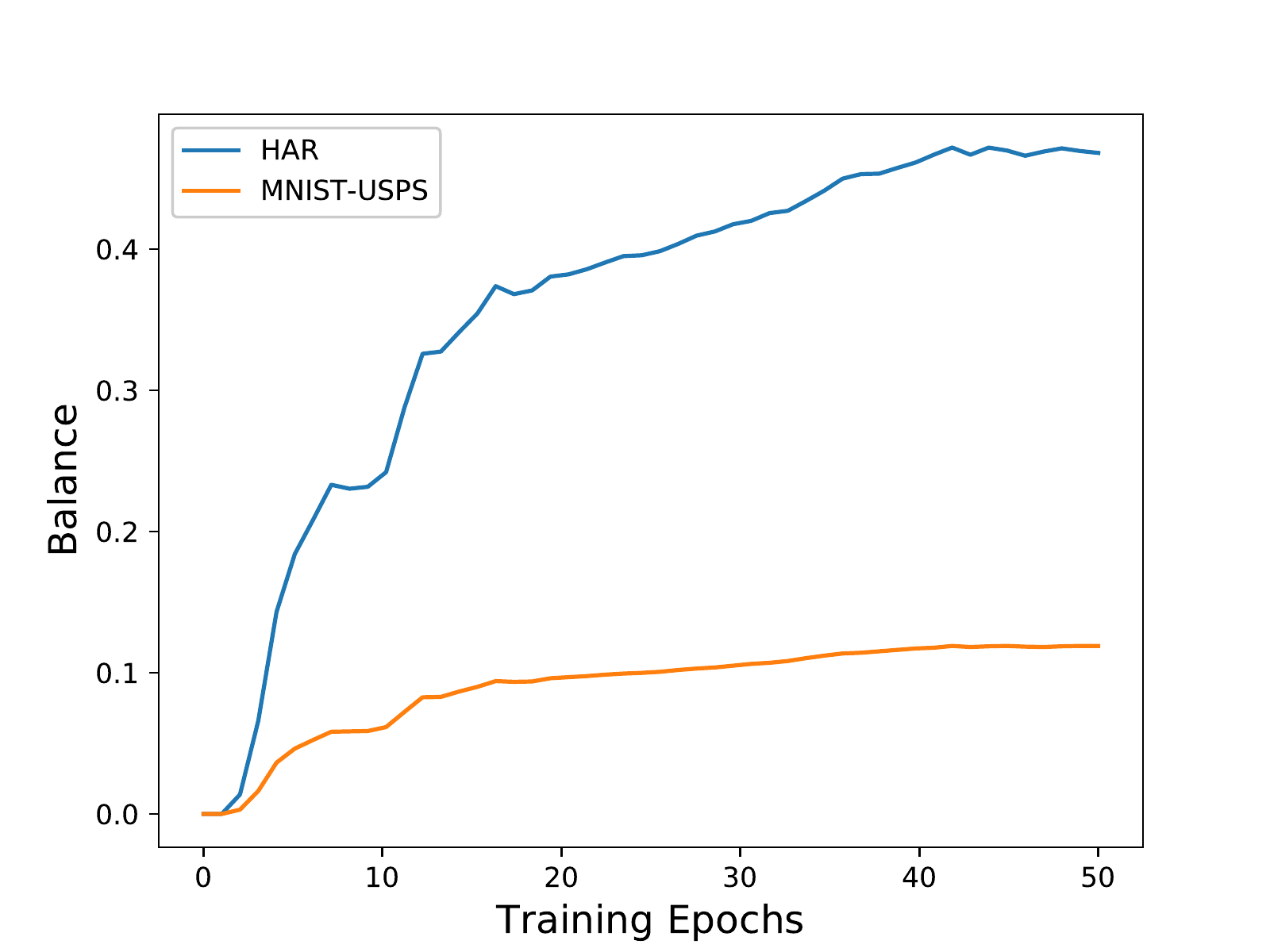}
 \captionof{figure}{Train Balance}
 \label{fig:convergence_balance}
\end{minipage}


\section{Conclusion}
In this paper, we explore the novel direction of adding fairness into deep clustering. This is a challenging problem given the end-to-end deep learning setting which does not facilitate pre-processing into fairlets and the need for scalability to large data sets. We formulate a group level measure of fairness as an integer linear programming problem and show the problem can be solved efficiently due to total unimodularity (Theorem \ref{th:fair}). We then add this solver into a discriminative deep learner and show that our formulation works with multi-state sensitive attributes as well as flexible fairness constraints that can occur in real-life applications.  Extensive experiments demonstrate the strong performance of our approach and an in-depth analysis including feature space visualization, hyper-parameter tuning, model convergence analysis, and investigating flexible fair constraints shows its versatility.

\bibliography{example_paper}
\bibliographystyle{abbrv}

\end{document}